\documentclass[letterpaper, 10 pt, conference]{ieeeconf}  

\IEEEoverridecommandlockouts                              

\usepackage{amsmath} 
\usepackage{amssymb}  
\usepackage{booktabs}
\usepackage{graphicx}
\usepackage[subrefformat=parens]{subcaption}
\usepackage{color}

\title{\LARGE \bf
3D Segmentation Using Viewpoint-Dependent Spatial Relationships
}

\author{
Ayaka Nanri$^{1,*}$,
Klara Reichard$^{2,3,*}$,
Mert Kiray$^{2,4,5,*}$,
Federico Tombari$^{2,6}$,
Benjamin Busam$^{2,4,5}$,
Asako Kanezaki$^{1,7,8}$\\[0.5em]
$^{1}$Institute of Science Tokyo, 
$^{2}$Technical University of Munich, 
$^{3}$BMW Group, \\
$^{4}$Munich Center for Machine Learning (MCML),
$^{5}$Obsphera, 
$^{6}$Google, 
$^{7}$Tohoku University,
$^{8}$RIKEN,\\
\small{$^{*}$Equal contribution}
}

\begin{document}

\maketitle
\thispagestyle{empty}
\pagestyle{empty}

\begin{abstract}

Recent advances in 3D datasets and multimodal models have greatly improved natural language 3D scene understanding. However, most 3D referring segmentation methods do not explicitly represent the observer viewpoint, making spatial relations such as ``left,'' ``right,'' ``front,'' and ``behind'' ambiguous and difficult to evaluate. We introduce a viewpoint-aware 3D referring segmentation dataset containing 220k benchmark samples, and scalable to tens of millions of viewpoint-conditioned samples through dense viewpoint sampling. In this dataset, target objects can only be identified through observer-centric spatial relations, making viewpoint-conditioned grounding necessary. We construct the benchmark by leveraging camera poses to automatically annotate observer-centric relations (left/right, front/behind) together with viewpoint-independent relations (above/under). Using this benchmark, we evaluate several existing 3D large multimodal models in a zero-shot setting and find that current models struggle with viewpoint-dependent spatial instructions. We further study how explicit viewpoint information can be incorporated into 3D large multimodal models. We introduce a viewpoint representation that encodes camera poses and conditions the model on the observation viewpoint, improving segmentation accuracy on viewpoint-dependent relations and increasing mIoU from 0.30 to 0.47 compared to a model without viewpoint conditioning. The dataset, code, and trained models will be made publicly available upon acceptance.

\end{abstract}

\section{Introduction}

Recent advances in 3D datasets and large multimodal models have greatly improved natural language interaction with 3D scenes, enabling applications in robotics, AR/VR, and autonomous systems. Among the core tasks in this area, \emph{3D referring segmentation} takes a natural language instruction and predicts the segmentation mask of the referred object in a 3D scene~\cite{chen2020scanrefer, zhang2023multi3drefer, ding2025multimodal}.

A key limitation of current approaches is that they rarely represent the observer viewpoint as an explicit input. This limitation is particularly problematic for observer-centric spatial relations such as \emph{left}, \emph{right}, \emph{front}, and \emph{behind}, whose meanings depend on the position and orientation of the observer. In real-world robotic manipulation and human--robot interaction, the ability to interpret object relations from the current viewpoint is essential. For example, instructions such as ``pick up the object to the right of the monitor'' cannot be interpreted without knowing the observation viewpoint.

Despite their importance, viewpoint-dependent relations are not rigorously evaluated in existing 3D referring segmentation benchmarks. Even when prompts include expressions such as left and right, the viewpoint is often implicit, and the target object can frequently be identified from object attributes alone. As a result, existing settings do not isolate viewpoint-dependent relational reasoning and provide limited insight into whether 3D large multimodal models actually use viewpoint information.

In this paper we study viewpoint-conditioned relational grounding in 3D and introduce a task called \emph{Viewpoint-Aware 3D Referring Segmentation}. The task explicitly provides the observation viewpoint and requires identifying the target object solely through spatial relations with respect to an anchor object. In conventional referring segmentation settings, relative spatial expressions may appear in language descriptions, but the viewpoint is typically not defined as part of the task formulation. This makes it difficult to evaluate viewpoint-dependent reasoning in a controlled manner. Furthermore, some prior works parameterize viewpoint information, but the target object can often still be uniquely identified from attributes such as object category or appearance, so viewpoint reasoning is not strictly required.

\begin{figure}[tb]
\centering

\begin{minipage}{0.49\linewidth}
    \centering
    \includegraphics[width=\linewidth]{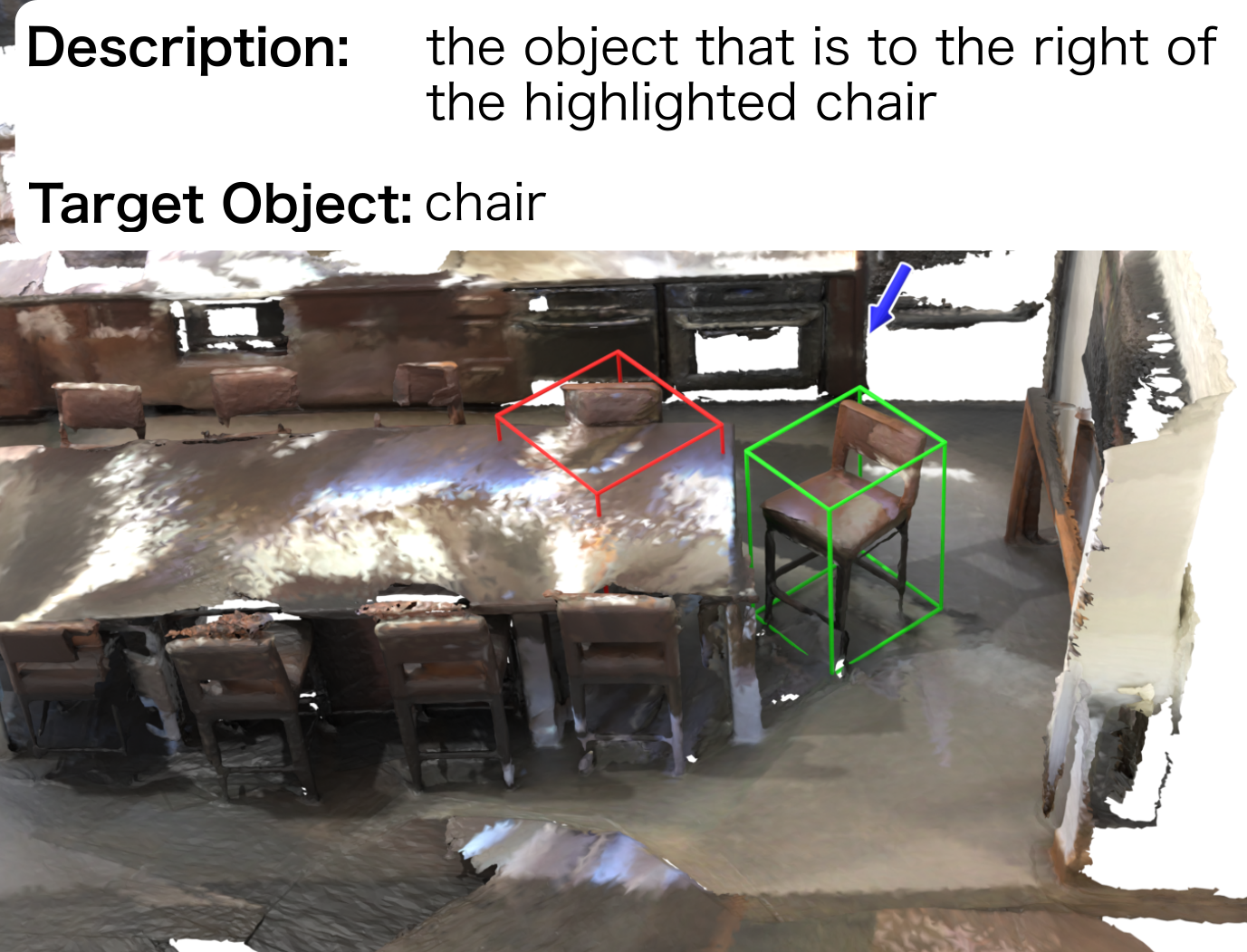}
    \subcaption{Ground Truth}
    \label{}
\end{minipage}
\hfill
\begin{minipage}{0.49\linewidth}
    \centering
    \includegraphics[width=\linewidth]{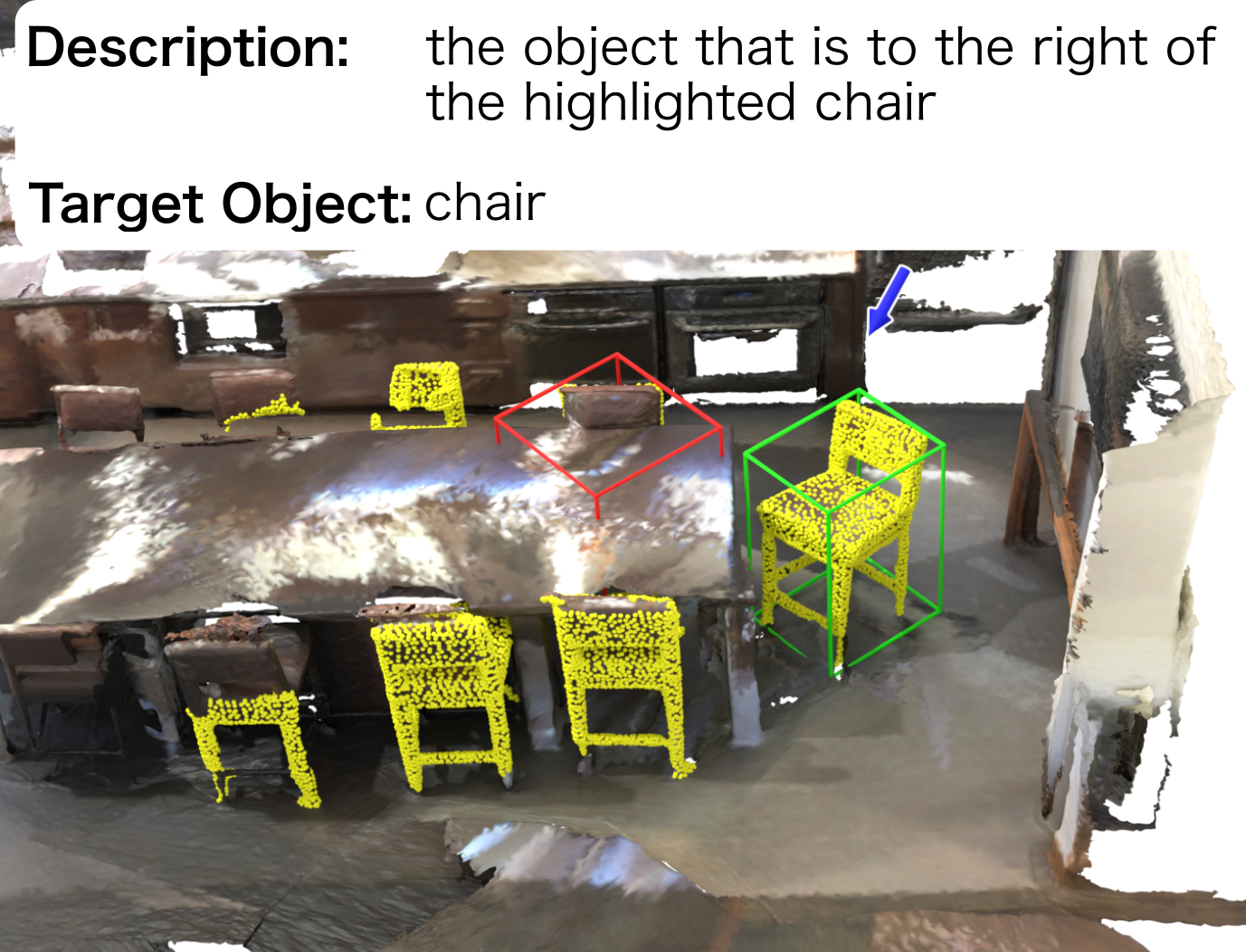}
    \subcaption{3D-LLaVA's prediction}
    \label{}
\end{minipage}

\vspace{0.5em}

\begin{minipage}{0.49\linewidth}
    \centering
    \includegraphics[width=\linewidth]{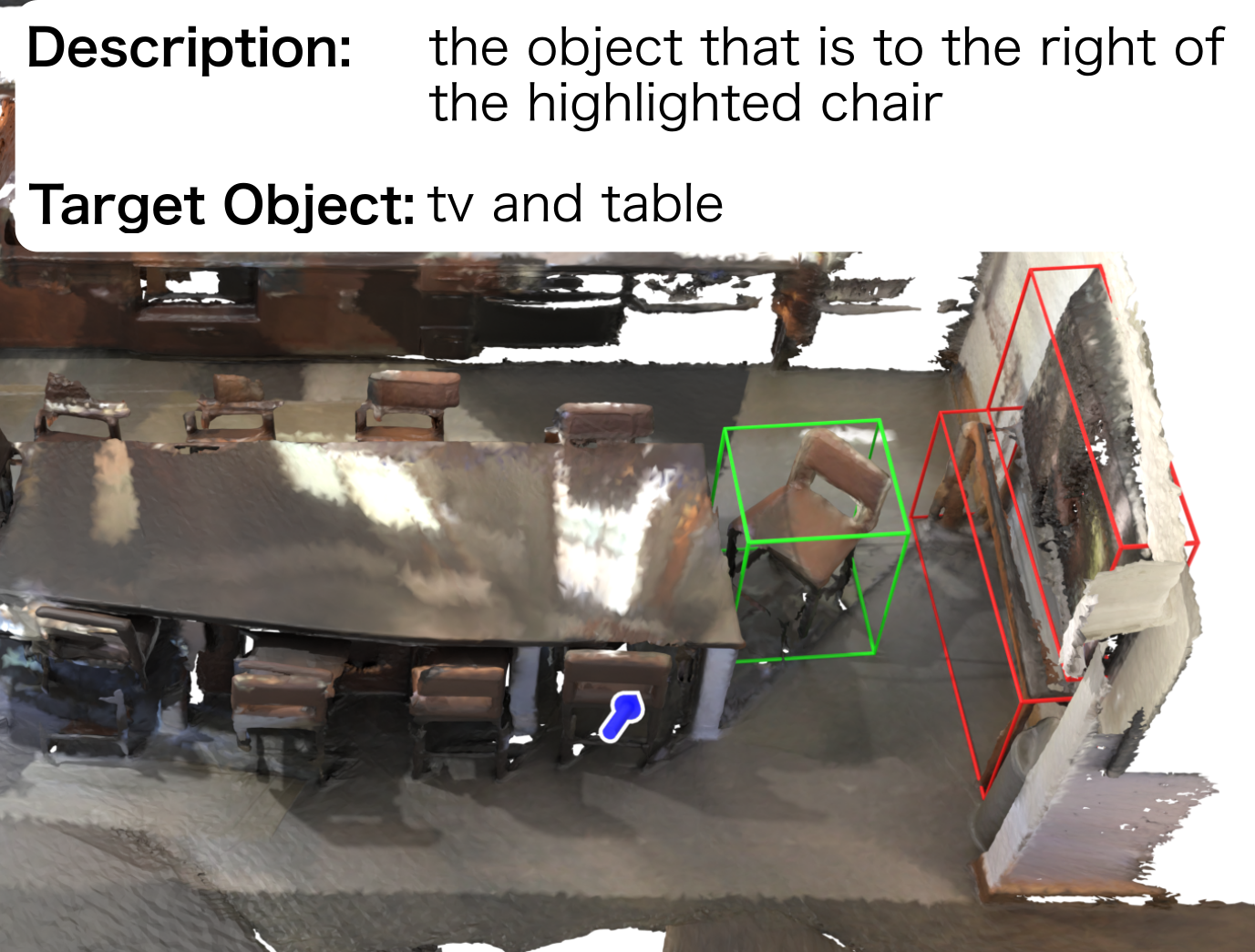}
    \subcaption{Ground Truth}
    \label{}
\end{minipage}
\hfill
\begin{minipage}{0.49\linewidth}
    \centering
    \includegraphics[width=\linewidth]{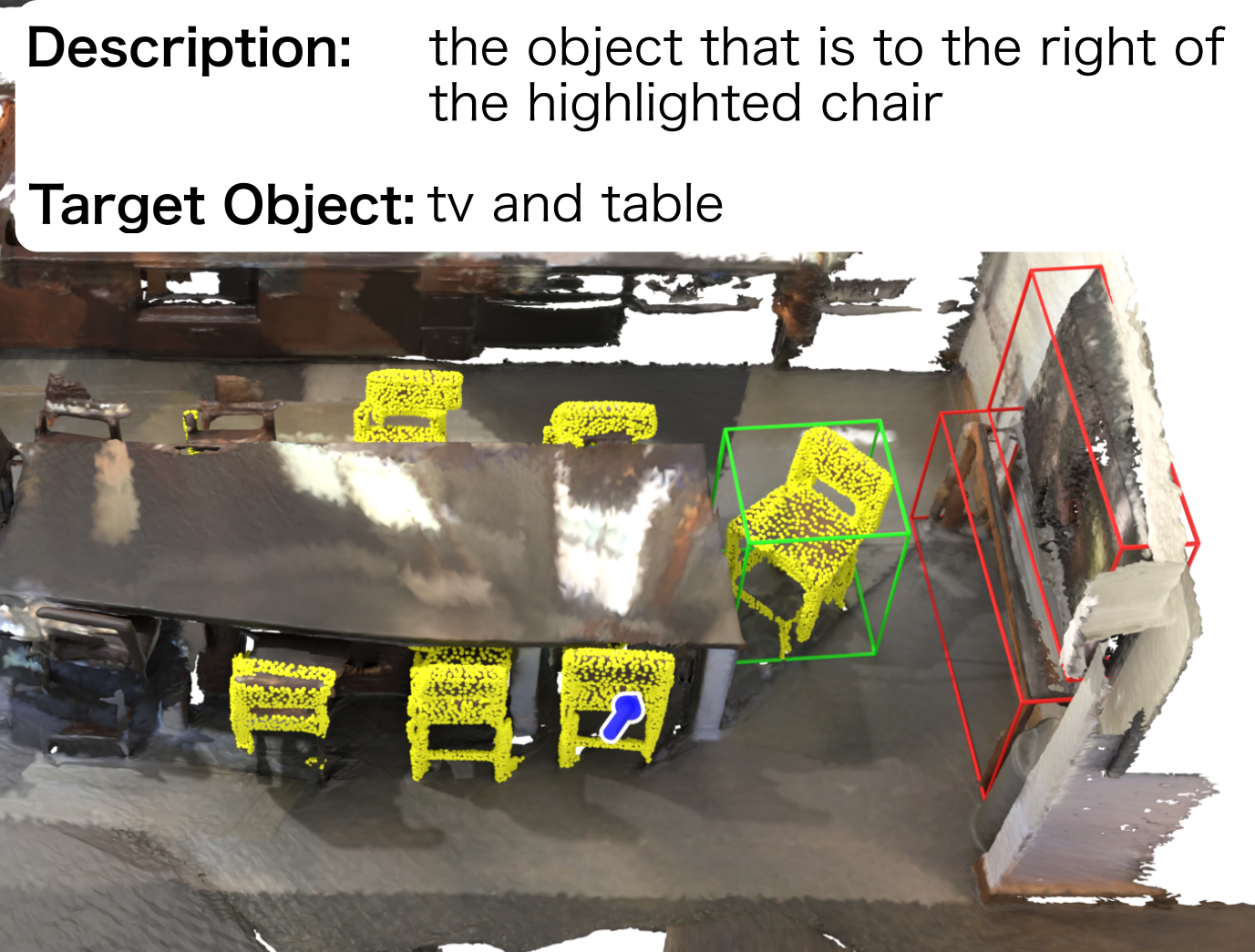}
    \subcaption{3D-LLaVA's prediction}
    \label{fig:panel_d}
\end{minipage}
\caption{\textbf{Limitation of existing 3D LMM in handling viewpoint-dependent spatial relationships.}
The same instruction is given in both scenes, but the correct target objects differ between (a) and (c) due to the change in viewpoint. However, 3D-LLaVA fails to account for the viewpoint and predicts multiple chairs in both cases (b, d).
Green bounding boxes indicate the anchor object, red bounding boxes denote the ground-truth target, blue arrows represent the viewpoint, and yellow points show the predicted mask.}
\label{fig:original_viewpoint_dependent}
\end{figure}
Figure~\ref{fig:original_viewpoint_dependent} illustrates a key limitation of current 3D large multimodal models (3D LMMs) in handling viewpoint-dependent spatial relationships. The same instruction refers to different target objects when the scene is observed from different viewpoints. However, 3D-LLaVA~\cite{deng20253dllava} fails to consider this viewpoint dependency.

To enable systematic evaluation of viewpoint-dependent spatial reasoning, we construct a dataset tailored to the proposed task and evaluate representative 3D large multimodal models. We further introduce an explicit viewpoint representation that encodes camera poses and integrates them into a 3D large multimodal model, and analyze how viewpoint information affects model behavior and performance.

\noindent\textbf{Our contributions are:}
\begin{itemize}

\item We introduce a viewpoint-aware 3D referring segmentation dataset, where target objects can only be identified through observer-centric spatial relations, making viewpoint-conditioned grounding strictly necessary.

\item We construct a large-scale benchmark with automatic annotation of observer-centric spatial relations using camera poses, enabling systematic evaluation of viewpoint-dependent reasoning.

\item We evaluate representative 3D large multimodal models on the proposed benchmark and introduce an explicit viewpoint representation that significantly improves segmentation performance on viewpoint-dependent spatial relations.

\end{itemize}

\section{Related Work}

\noindent \textbf{Inter-Object Relations in 3D Scenes.}
Inter-object relations in 3D scenes describe spatial relationships between multiple objects rather than attributes of a single object. These include distance relations (e.g., near, far), containment and contact relations (e.g., inside, on top of, touching), and relative positional relations (e.g., left, right, front, behind, above, below). In computer vision, inter-object relations are commonly formulated as predicates between object pairs, such as in visual relationship detection~\cite{lu2016visual} and scene graph representations~\cite{wald2020learning}. The meaning of spatial relations is defined with respect to a reference frame, such as world-centered, object-centered, or observer-centered coordinates~\cite{levinson2003space}. Relative positional relations depend on the observer's viewpoint. For example, the relation ``A is to the left of B'' may reverse with changes in position or orientation, making such relations ambiguous without an explicit viewpoint.

\noindent \textbf{Datasets with Inter-Object Relations without Explicit Viewpoint.}
With recent progress in integrating language and 3D spatial understanding, many datasets targeting 3D spatial reasoning have been proposed~\cite{SegPoint, ReferSplat, wu20243d, yin2025spatial, yang20253d}. ScanQA~\cite{azuma_2022_CVPR}  focuses on question answering in 3D scenes, while Scan2Cap~\cite{chen2021scan2cap} addresses object captioning. Although some prompts in these datasets involve inter-object relations, viewpoint information is not explicitly specified. As a result, such relations are typically interpreted either in an object-centered manner or according to annotator assumptions.

For language-guided segmentation, representative datasets include ScanRefer~\cite{chen2020scanrefer} and Multi3DRefer~\cite{zhang2023multi3drefer}. ScanRefer~\cite{chen2020scanrefer} defines a task to localize a target object in a 3D scene using natural language descriptions, while Multi3DRefer~\cite{zhang2023multi3drefer} considers more realistic scenarios involving multiple target objects. Nr3D~\cite{achlioptas2020referit_3d} explicitly handles object-centered relative relations. However, these datasets do not explicitly define observer viewpoints and therefore cannot rigorously evaluate viewpoint-dependent relations.

\noindent \textbf{Datasets Handling Viewpoint-Dependent Relations.}
Several datasets attempt to incorporate viewpoint information into spatial reasoning tasks. SQA3D~\cite{ma2022sqa3d} describes the agent's position and orientation in natural language and requires the model to interpret the environment before answering questions. SURPRISE3D~\cite{huang2025surprise3d} introduces several spatial reasoning tasks with parametric viewpoint descriptions. However, prompts may allow the target object to be identified from attribute information alone, reducing the need for viewpoint-dependent reasoning. RoboSpatial~\cite{song2025robospatial} addresses observer-centered, object-centered, and world-centered relations, but its target task is identifying free space rather than instance segmentation.

Overall, existing datasets do not provide a controlled setting where viewpoint-dependent relations must be resolved using explicit camera poses while removing attribute-based shortcuts. This limitation motivates the viewpoint-aware benchmark proposed in this work.

\noindent \textbf{3D Large Multimodal Models.}
With the rapid development of large language models (LLMs) and multimodal learning, numerous studies have extended these frameworks to handle 3D information~\cite{zhang2021pointclip, man2024situation, tang2024minigpt, SpatialLM, Chen_2024_CVPR}. 3D large multimodal models (3D LMMs) aim to perform tasks such as question answering, caption generation, and segmentation over 3D scenes using unified architectures.

Models such as ChatScene~\cite{huang2024chat}, 3D-LLaVA~\cite{deng20253dllava}, LL3DA~\cite{chen2024ll3da}, and Reason3D~\cite{reason3d} directly process point clouds, while 3D-LLM~\cite{3dllm} and LLaVA-3D~\cite{zhu2024llava} integrate multi-view images or RGB-D observations. Although these models achieve strong performance on object recognition and general spatial reasoning tasks, they do not treat viewpoint as an explicit conditioning variable. Consequently, viewpoint-dependent spatial relations are often interpreted based on implicit coordinate assumptions or statistical biases in the training data, leading to inconsistent interpretations across viewpoints.

\section{Dataset Construction}

\subsection{Dataset Design Policy}

We construct the \emph{Viewpoint-Aware 3D Referring Segmentation Benchmark} to evaluate models that can correctly understand and reason about viewpoint-dependent three-dimensional spatial relationships.
Our dataset is built upon ScanNet~\cite{dai2017scannet}, a large-scale 3D dataset of real-world indoor environments. ScanNet provides RGB-D image sequences, corresponding camera poses, reconstructed 3D scenes, and instance-level object annotations. It also defines 50 object categories, mainly consisting of furniture and everyday items, covering diverse object arrangements and spatial structures in real environments.
Existing 3D referring expression datasets, such as ScanRefer~\cite{chen2020scanrefer} and Multi3DRefer~\cite{zhang2023multi3drefer}, primarily focus on viewpoint-independent object identification and do not strictly handle observer-dependent relations such as ``to the left of'' or ``to the right of.'' Therefore, in this work, we design a dataset centered on spatial relationships defined with respect to the observer’s viewpoint.

We consider the following six spatial relations:
\texttt{left}, \texttt{right}, \texttt{front}, \texttt{behind}, \texttt{above}, and \texttt{under}.
Among them, \texttt{left}, \texttt{right}, \texttt{front}, and \texttt{behind} are viewpoint-dependent relations, whereas \texttt{above} and \texttt{under} are viewpoint-independent relations defined based on the gravity direction.
Furthermore, the objects involved in the relations are restricted to instance-level annotations provided in ScanNet~\cite{dai2017scannet}, and background elements such as walls, floors, and ceilings are excluded. This design allows us to focus on meaningful object-to-object relationships.

\subsection{Definition of Coordinate Systems}

We employ two reference frames to define spatial relationships: the camera coordinate system and the world coordinate system. To avoid ambiguity, we denote the camera coordinate system as $(X_{\mathrm{cam}}, Y_{\mathrm{cam}}, Z_{\mathrm{cam}})$ and the world coordinate system as $(X_{\mathrm{W}}, Y_{\mathrm{W}}, Z_{\mathrm{W}})$.

The camera coordinate system $(X_{\mathrm{cam}}, Y_{\mathrm{cam}}, Z_{\mathrm{cam}})$ is defined for each observation viewpoint. Its origin is located at the camera center; the $Z_{\mathrm{cam}}$ axis corresponds to the viewing direction; the $X_{\mathrm{cam}}$ axis corresponds to the horizontal direction of the image plane; and the $Y_{\mathrm{cam}}$ axis corresponds to the vertical direction. This forms a right-handed coordinate system. In this coordinate system, left/right and front/behind relationships are defined.

In contrast, the world coordinate system $(X_{\mathrm{W}}, Y_{\mathrm{W}}, Z_{\mathrm{W}})$ is fixed to the entire scene, where the $Y_{\mathrm{W}}$ axis represents the gravity direction. In this work, the world coordinate system is used only to determine vertical relationships (above / under).

\begin{figure*}[tb]
  \centering
  \includegraphics[width=.9\linewidth]
  {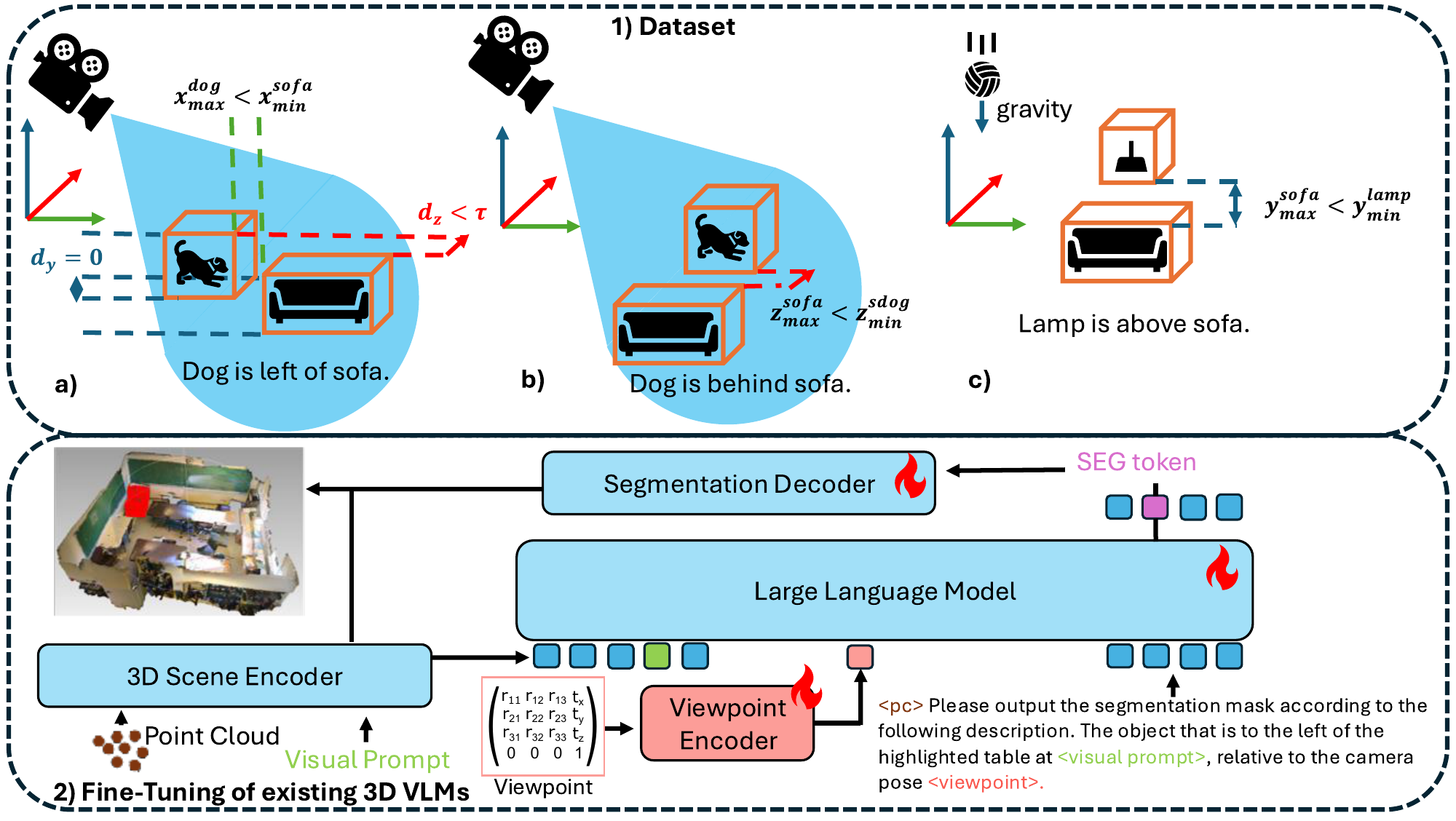}
  \caption{Dataset generation and finetuning of the viewpoint-conditioned 3D large multimodal model.}
  \label{fig:model_structure}
\end{figure*}

\subsection{Automatic Annotation Method}

The proposed dataset is generated fully automatically using the 3D point clouds and RGB-D sequences included in ScanNet~\cite{dai2017scannet}, along with their associated camera pose information. Figure~\ref{fig:model_structure}~(dataset a–c) provides a visual overview of the automatic annotation pipeline.
For each scene, multiple camera poses are sampled, and each is treated as an observation viewpoint. At each viewpoint, visible object instances are extracted using the camera frustum, and background classes (walls, floors, and ceilings) are excluded.
For every pair of visible objects $(o_i, o_j)$, spatial relationships are determined using their 3D bounding boxes. In this study, we do not use the bounding box centers; instead, relationships are defined by strict geometric conditions based on the minimum and maximum values along each axis.
For each object $o$, we denote its bounding box in the camera coordinate system as
$
[x_{\min}^{o,\mathrm{cam}}, x_{\max}^{o,\mathrm{cam}}]$
and
$
[z_{\min}^{o,\mathrm{cam}}, z_{\max}^{o,\mathrm{cam}}],
$
and its vertical range in the world coordinate system as
$
[y_{\min}^{o,\mathrm{W}}, y_{\max}^{o,\mathrm{W}}].
$

\paragraph{Left / Right (Viewpoint-Dependent)}

The left/right relationship between $o_i$ and $o_j$ is defined based on separation along the $X_{\mathrm{cam}}$ axis and overlap constraints along the other two axes:
\[
o_i \footnotesize{\text{ is left of }} o_j
\Leftrightarrow
x_{\max}^{o_i, \mathrm{cam}} \! < \! x_{\min}^{o_j, \mathrm{cam}}
\land
d_y(i,j) \! < \! \tau
\land
d_z(i,j) \! < \! \tau
\]
\[
o_i \footnotesize{\text{ is right of }} o_j
\Leftrightarrow
x_{\min}^{o_i, \mathrm{cam}} \! > \! x_{\max}^{o_j, \mathrm{cam}}
\land
d_y(i,j) \! < \! \tau
\land
d_z(i,j) \! < \! \tau
\]
Here, $d_y(i,j)$ and $d_z(i,j)$ denote the minimum distances between the two bounding boxes along the $y$- and $z$-axes, respectively, and $\tau$ is the maximum allowed misalignment (set to $0.5$ in this study). These conditions ensure that the left/right relation is assigned only when the objects are approximately aligned along the other two axes.

\paragraph{Front / Behind (Viewpoint-Dependent)}

Similarly, depth relationships are defined based on the $Z_{\mathrm{cam}}$ axis:
\[
o_i \footnotesize{\text{ is in front of }} o_j
 \! \Leftrightarrow \!
z_{\max}^{o_i, \mathrm{cam}} \! < \! z_{\min}^{o_j, \mathrm{cam}}
\land
d_x(i,j) \! < \! \tau
\land
d_y(i,j) \! < \! \tau
\]
\[
o_i \footnotesize{\text{ is behind }} o_j
\Leftrightarrow
z_{\min}^{o_i, \mathrm{cam}} \! > \! z_{\max}^{o_j, \mathrm{cam}}
\land
d_x(i,j) \! < \! \tau
\land
d_y(i,j) \! < \! \tau
\]

\paragraph{Above / Under (Viewpoint-Independent)}

Vertical relationships are defined based on the $Y_{\mathrm{W}}$ axis of the world coordinate system:
\[
o_i \footnotesize{\text{ is under }} o_j
\Leftrightarrow
y_{\max}^{o_i, \mathrm{W}} \! < \! y_{\min}^{o_j, \mathrm{W}}
\land
d_x(i,j) \! < \! \tau
\land
d_z(i,j) \! < \! \tau
\]
\[
o_i \footnotesize{\text{ is above }} o_j
\Leftrightarrow
y_{\min}^{o_i, \mathrm{W}} \! > \! y_{\max}^{o_j, \mathrm{W}}
\land
d_x(i,j) \! < \! \tau
\land
d_z(i,j) \! < \! \tau
\]

\paragraph{Sample Construction}

In this dataset, each data sample is reconstructed as a quadruple:
$
(\text{scene}, \text{viewpoint}, \text{anchor object}, \text{relation}),
$
and the corresponding target objects are assigned as ground-truth labels.
Here, \texttt{scene} denotes each 3D indoor scan included in ScanNet. Although ScanNet contains 707 distinct indoor environments, each environment is scanned multiple times. In this work, all 1,513 scans are treated as independent scenes.
The \texttt{viewpoint} corresponds to a camera pose at a given observation, represented as a $4\times4$ homogeneous transformation matrix including rotation and translation.
The \texttt{anchor object} is the reference object for defining the spatial relationship. For each viewpoint, all visible object instances excluding background classes are considered as anchor candidates.
The \texttt{relation} represents one of the six spatial relationships defined in the previous section or their combinations.

Based on the geometric conditions defined above, spatial relationships are determined independently for each visible object pair $(o_i, o_j)$ at each viewpoint. Since each relation is evaluated independently, an object pair may simultaneously satisfy multiple relations, such as ``left'' and ``front.'' All satisfied relations are enumerated and used for annotation. Objects that are occluded from the viewpoint are excluded.

The final dataset is reconstructed not at the object-pair level, but at the quadruple level described above. Specifically, within the same scene and viewpoint, for a given anchor object $o_j$ and a specific relation (e.g., ``right''), all objects $o_i$ satisfying that relation are grouped together and assigned as the target objects. Therefore, a single referring expression may correspond to multiple target instances rather than a single object.
Each data sample finally contains the following information:
\begin{itemize}
    \item 3D point cloud of the scene
    \item Observation viewpoint (camera pose)
    \item Relation (e.g., ``right'', ``behind'')
    \item Segmentation mask of the anchor object
    \item Segmentation masks of the target objects (Ground Truth)
\end{itemize}
Here, the anchor object serves as the reference of the relation, and the target objects are the ground-truth objects specified by that relation.
\begin{figure*}[tb]
    \begin{tabular}{cc}
        \begin{minipage}[t]{0.24\hsize}
        \centering
        \includegraphics[keepaspectratio, width=1.0\linewidth]{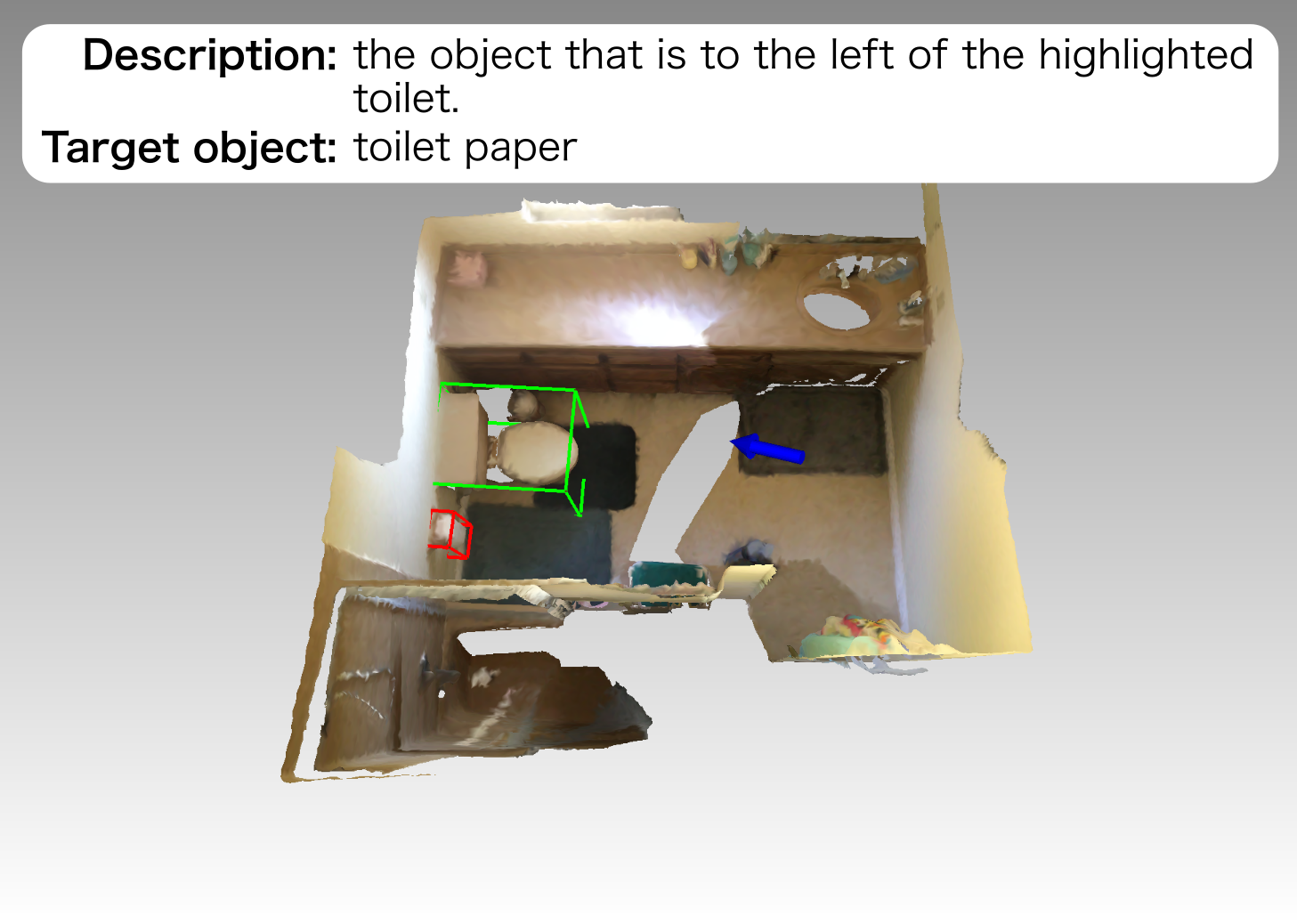}
        \end{minipage} 
        \begin{minipage}[t]{0.24\hsize}
        \centering
        \includegraphics[keepaspectratio, width=1.0\linewidth]{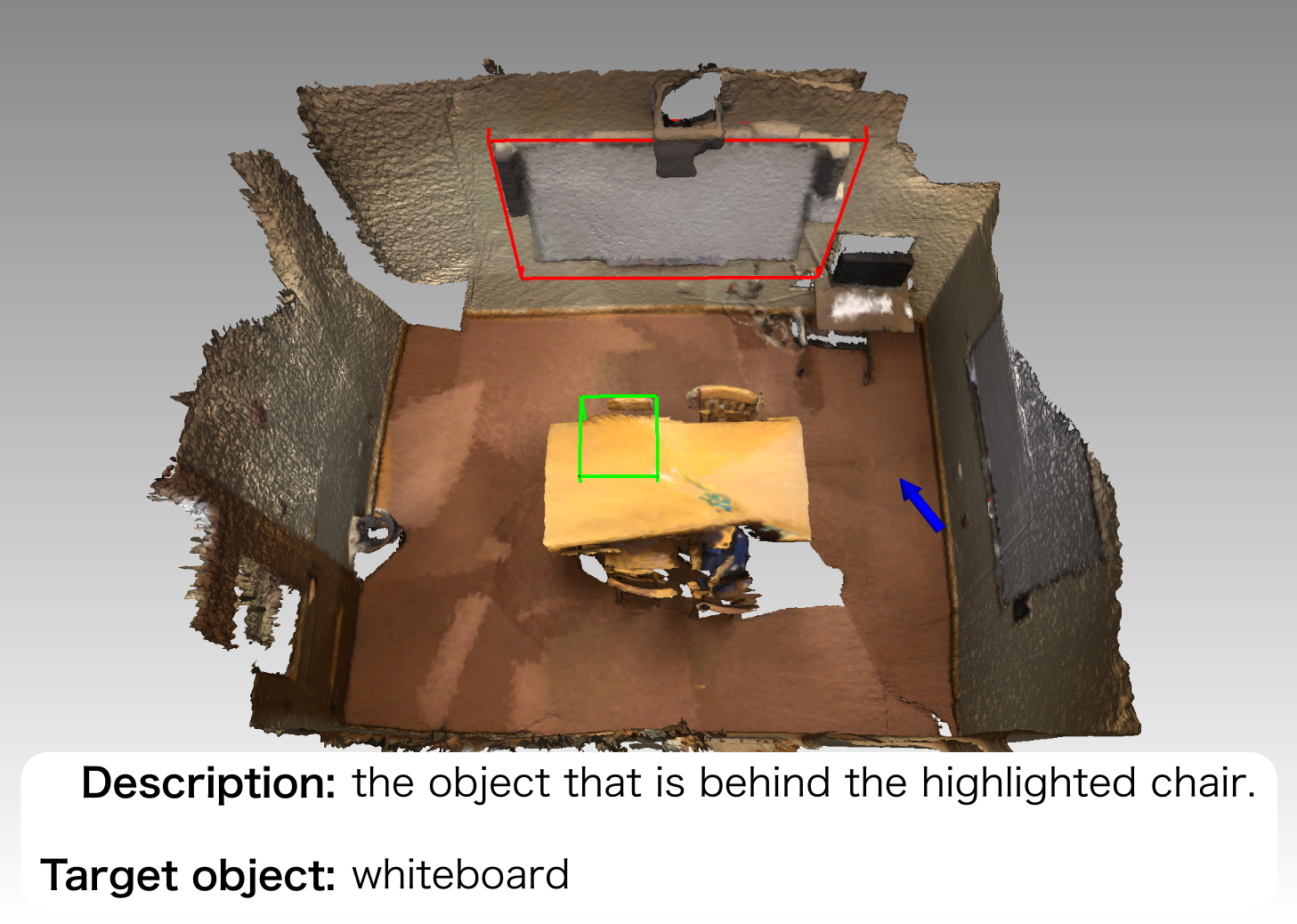}
        \end{minipage} 
        
        \begin{minipage}[t]{0.24\hsize}
        \centering
        \includegraphics[keepaspectratio, width=1.0\linewidth]{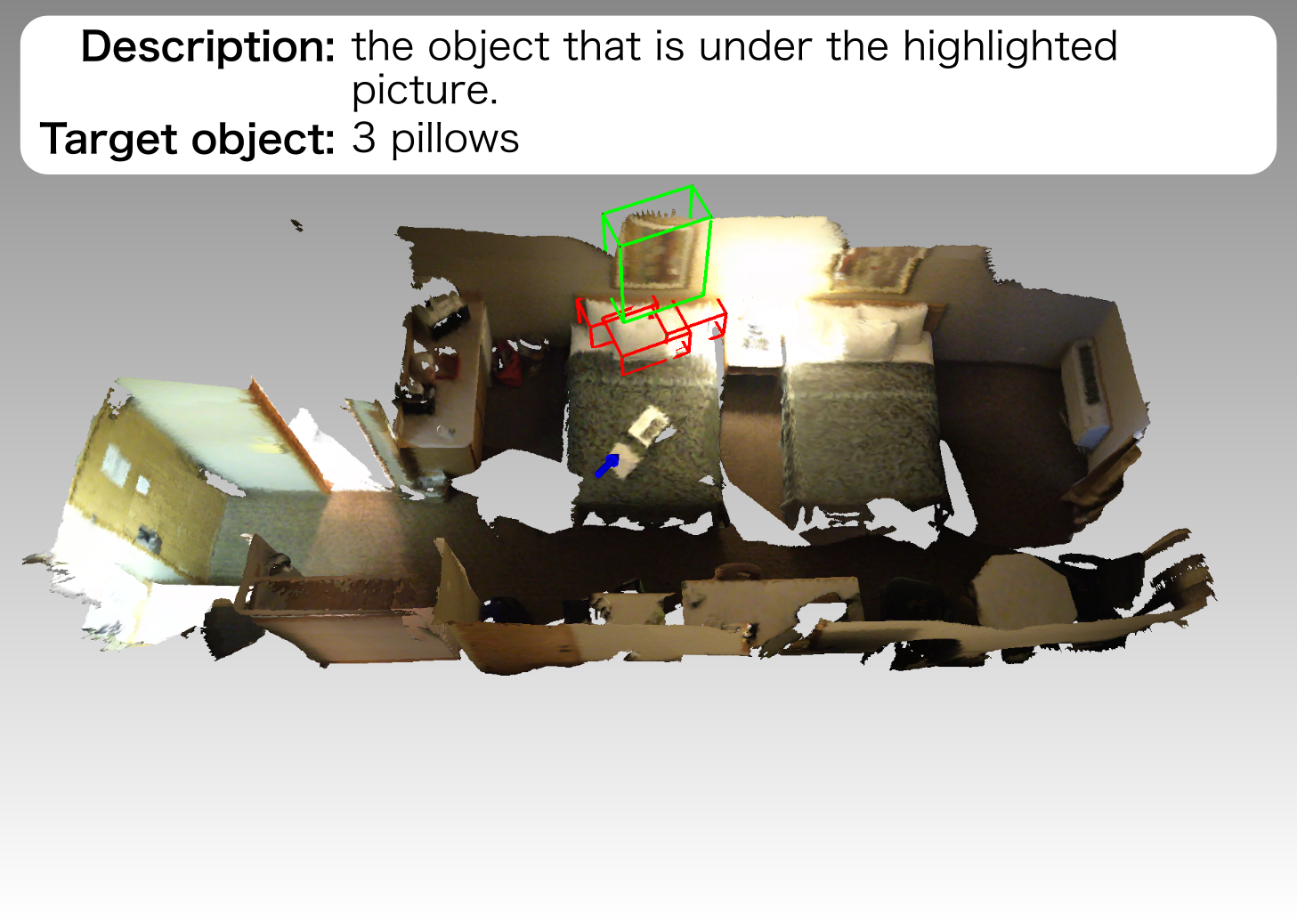}
        \end{minipage} 
        \begin{minipage}[t]{0.24\hsize}
        \centering
        \includegraphics[keepaspectratio, width=1.0\linewidth]{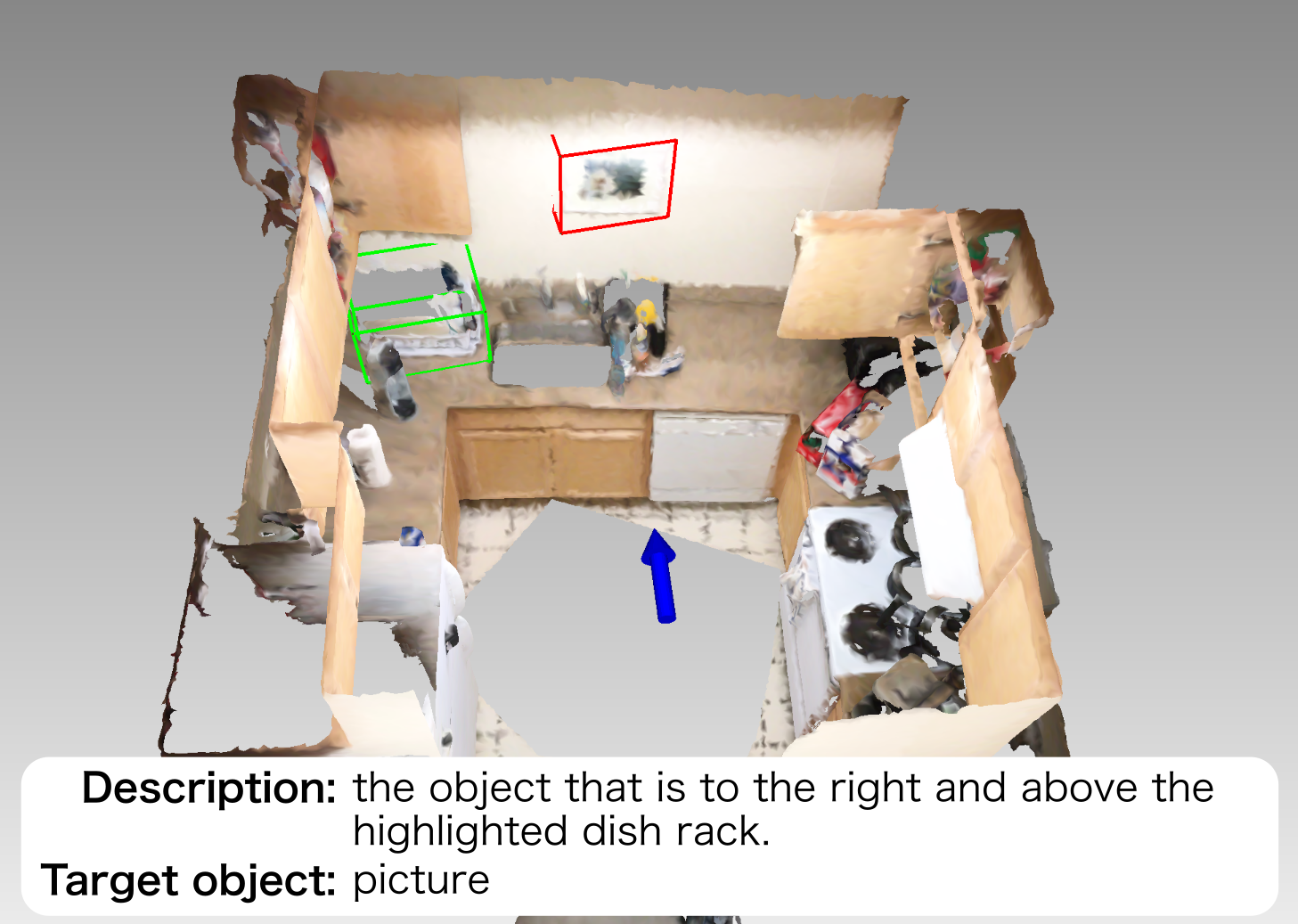}
        \end{minipage} 
    \end{tabular}
    \caption{\textbf{Examples of viewpoint-aware 3D referring segmentation samples.} The anchor object is shown with a green bounding box, target objects are shown with red bounding boxes, and the observation viewpoint is indicated by a blue arrow.}
    \label{fig:dataset_examples}
\end{figure*}
Figure~\ref{fig:dataset_examples} presents representative samples of the dataset, showing the anchor object and the corresponding target objects defined by the specified spatial relation.
\subsection{Dataset Statistics}

We constructed a dataset using 500 viewpoints per scene.  
However, under this setting, the total number of samples reaches approximately 10.8 million.  
Therefore, considering computational resources and experimental efficiency, we additionally created a lightweight version of the dataset using only 10 viewpoints per scene.
The spatial relations included in the dataset consist of six basic relations:
\texttt{left}, \texttt{right}, \texttt{front}, \texttt{behind}, 
\texttt{above}, and \texttt{under},  
as well as their combinations.  
Their frequencies are not uniformly distributed due to scene layouts and viewpoint distributions. 

\begin{table}[t]
\centering
\caption{Number of samples for each spatial relation}
\label{tab:relation_stats}
\begin{tabular}{lcc}
\toprule
Relation & 10 viewpoint/scene & 500 viewpoint/scene\\
\midrule
left   & 44,397 & 2,175,608\\
right  & 45,132 & 2,179,326\\
front  & 24,488 & 1,202,963\\
behind & 22,064 & 1,083,996\\
above  & 17,268 & 852,581\\
under  & 14,146 & 696,583\\
left, front   & 6,064 & 298,727\\
left, behind  & 5,719 & 286,702\\
left, above  & 3,491 & 169,993\\
left, under & 3,539 & 172,753\\
right, front  & 5,755 & 289,332\\
right, behind  & 6,048 & 296,335\\
right, above  & 3,562 & 175,289\\
right, under  & 3,450 & 166,769\\
front, above   & 1,700 & 81,399\\
front, under  & 2,241 & 106,579\\
behind, above  & 2,215 & 105,628\\
behind, under & 1,795 & 83,959\\
left, front, above & 1,068 & 53,144\\
left, front, under & 878 & 42,336\\
left, behind, above & 808 & 39,869\\
left, behind, under & 1,120 & 53,651\\
right, front, above & 1,050 & 51,599\\
right, front, under & 824 & 41,357\\
right, behind, above & 870 & 41,346\\
right, behind, under & 1,149 & 54,728\\
\midrule
Total   & 220,841 & 10,802,552\\
\bottomrule
\end{tabular}
\end{table}

We therefore count the number of referring expression samples corresponding to each relation and summarize them in Table~\ref{tab:relation_stats}.  
Here, one sample refers to a single referring segmentation problem defined by a tuple of (scene, viewpoint, anchor object, relation). 
%
Both single and composite relations are represented at scale, with approximately balanced complementary pairs such as \texttt{left/right} and \texttt{front/behind}. 

%

We further compare the scale of our dataset with existing 3D referring expression datasets in Table~\ref{tab:dataset_stats}. The proposed dataset significantly surpasses representative prior datasets in the number of samples explicitly involving inter-object relationships, demonstrating that it provides large-scale and dense relation annotations.

\begin{table}[t]
\centering
\caption{Number of samples with inter-object relations in prompts in each dataset}
\label{tab:dataset_stats}
\begin{tabular}{lc}
\toprule
Dataset & \# Samples\\
\midrule
ScanRefer & 26,369\\
Multi3DRefer & 27,774\\
Nr3D & 10,952\\
SURPRISE3D & 50,466\\
Proposed Dataset (10 viewpoints/scene) & 220,841\\
Proposed Dataset (500 viewpoints/scene) & 10,802,552\\
\bottomrule
\end{tabular}
\end{table}

\subsection{Task Definition}

The task in this work is defined as \emph{Viewpoint-Aware 3D Referring Segmentation}.
The input consists of:
\begin{itemize}
    \item A 3D point cloud $\mathcal{P}$ (ScanNet scene)
    \item A camera pose $T \in SE(3)$ representing the observation viewpoint
    \item A segmentation mask of the anchor object $M_{\mathrm{anc}}$
    \item A natural language expression $q$ containing a spatial relation  
    (e.g., ``the object that is to the right of the highlighted bed at \texttt{<loc>}, relative to the camera pose \texttt{<viewpoint>}.'' )
\end{itemize}
The natural language expression $q$ explicitly includes the anchor object and one or more spatial relations (\texttt{left}, \texttt{right}, \texttt{front}, \texttt{behind}, \texttt{above}, \texttt{under}, etc.), and indirectly specifies the target object based on the observation viewpoint.  
For example:
``the object that is to the right of the highlighted bed at \texttt{<loc>}, relative to the camera pose \texttt{<viewpoint>}''.
In this task, referring expressions adopt a formulation that \emph{specifies the target object solely through spatial relations}, such as
``the object that is to the right of the bed.''
Other visual or semantic cues such as color, shape, or function are intentionally excluded. Therefore, identifying the target object necessarily requires interpreting the spatial relation conditioned on both the anchor object and the observation viewpoint.

The model output is the \emph{3D segmentation mask of the target objects}, denoted as $M_{\mathrm{tgt}}$, determined based on the specified relation.  
The target may consist of multiple objects, defined as the set of all instances satisfying the relation with respect to the anchor object.
Under this task formulation, the model must simultaneously demonstrate:
\begin{itemize}
    \item Understanding of 3D geometric structure
    \item Interpretation of viewpoint-dependent spatial relations
    \item Grounding of natural language references
\end{itemize}
In particular, even within the same scene and for the same object pair, relations such as \texttt{left} and \texttt{right} may invert when the viewpoint changes. Thus, this benchmark rigorously evaluates whether the model explicitly utilizes viewpoint information for spatial reasoning.


\section{Extending 3D Large Multimodal Models to Incorporate Viewpoints}

\subsection{Reference 3D Large Multimodal Architecture}

We adopt 3D-LLaVA~\cite{deng20253dllava} as a reference 3D large multimodal model to instantiate and evaluate the proposed viewpoint encoder.
3D-LLaVA is a general-purpose 3D large multimodal model that integrates a large language model with 3D visual representations, enabling tasks such as question answering, dense captioning, and referring segmentation within a unified architecture.


Given a 3D point cloud where each point contains spatial coordinates (XYZ) and color information (RGB), 3D-LLaVA encodes a point cloud into visual tokens using a sparse 3D backbone with superpoint-based aggregation and processes them jointly with language tokens through a large language model.

Visual prompts such as clicks or masks are embedded into the same feature space as the superpoint tokens, enabling interactive grounding. 
For referring segmentation, the hidden state of the special token \texttt{[SEG]} is used to decode the target 3D segmentation mask.

However, existing 3D large multimodal models, including 3D-LLaVA, do not explicitly incorporate observer viewpoint information. 
As a result, viewpoint-dependent spatial relations such as \texttt{left} and \texttt{right} must be inferred implicitly, often leading to inconsistent interpretations.

\subsection{Viewpoint Encoder}

We propose a model-agnostic viewpoint encoder that enables explicit viewpoint-conditioned reasoning in 3D large multimodal models.
The encoder is designed to be compatible with architectures that jointly process 3D visual tokens and language tokens, without requiring changes to the core 3D perception or language modules.

Each observation viewpoint is represented by a camera pose $T \in SE(3)$,
consisting of a rotation matrix $R \in SO(3)$ and a translation vector
$t \in \mathbb{R}^3$, represented as a $4 \times 4$ homogeneous transformation matrix.
The matrix is flattened into a vector and processed by a multilayer perceptron to produce a viewpoint embedding. The resulting embedding is treated as a viewpoint token and concatenated with the 3D visual tokens and language tokens before being provided as input to the language model.
This design allows viewpoint information to be incorporated in a unified token-based representation, consistent with existing multimodal transformer architectures. With explicit viewpoint conditioning, spatial expressions such as ``the object to the right of the bed'' are interpreted as geometric relations relative to the observation viewpoint, rather than being inferred implicitly from language or dataset biases.
A generalized architecture illustrating how the viewpoint encoder integrates with 3D large multimodal models is shown in Fig.~\ref{fig:model_structure}~(2).

In this work, we instantiate the proposed viewpoint encoder on 3D-LLaVA~\cite{deng20253dllava}, but the design is not specific to this model and can be readily applied to other 3D large multimodal architectures. In addition to 3D-LLaVA, we also integrate the proposed viewpoint encoder into Reason3D~\cite{reason3d}, demonstrating its applicability across different 3D multimodal reasoning architectures.




\subsection{Training Setup}

We initialize the model with pretrained 3D-LLaVA weights and adapt it to viewpoint-aware referring segmentation.

During training, we update only:

\begin{itemize}
    \item the viewpoint encoder,
    \item LoRA parameters inserted into the language model,
    \item the visual projection layer and modules associated with the \texttt{[SEG]} token,
\end{itemize}

while keeping the Sparse U-Net, OST, and the main language model weights frozen.

Low-Rank Adaptation (LoRA) enables efficient fine-tuning while keeping the majority of the model parameters fixed. This design allows the model to incorporate viewpoint information without retraining the full network.

Training uses a multitask objective that jointly optimizes language generation and segmentation:

\[
\mathcal{L} =
\mathcal{L}_{\text{text}}
+
\lambda \, \mathcal{L}_{\text{mask}}.
\]

Here $\mathcal{L}_{\text{text}}$ denotes the cross-entropy loss for next-token prediction, and $\mathcal{L}_{\text{mask}}$ denotes the segmentation loss composed of binary cross-entropy and Dice loss.s

\section{Experiments}

\subsection{Experimental Setup}
Unless otherwise stated, we train models on the lightweight split of our dataset with 10 viewpoints per scene and evaluate on the full split with 500 viewpoints per scene. We report mean Intersection-over-Union (mIoU) and Acc@0.25 and Acc@0.50, which measure the fraction of predictions whose IoU exceeds 0.25 and 0.50, respectively.
When multiple ground-truth target instances exist for a query, we compute IoU between the prediction and each individual target instance as well as the union mask of all target instances, and report the maximum IoU. This protocol evaluates whether the model follows the intended relational direction, without requiring perfect segmentation of all valid target instances.

For fine-tuning experiments, we optimize a multitask objective consisting of a text generation loss and a segmentation loss:
\[
\mathcal{L}=\mathcal{L}_{\text{text}}+\lambda\,\mathcal{L}_{\text{mask}}.
\]
Here, $\mathcal{L}_{\text{text}}$ is the cross-entropy loss for next-token prediction, and $\mathcal{L}_{\text{mask}}$ is composed of binary cross-entropy and Dice loss. Unless otherwise noted, we set $\lambda=0.1$.

We fine-tune the model by updating only the viewpoint encoder, the LoRA parameters inserted into the LLM, and the projection and segmentation-related modules (including those associated with \texttt{[SEG]}), while keeping the Sparse U-Net and OST frozen. Unless otherwise specified, we train for 1 epoch with viewpoint encoder hidden dimension set to 4096. All other training hyperparameters (optimizer, learning rate, batch size, and LoRA configuration) are fixed across experiments and reported in the supplementary material.

During preliminary experiments, we observed unstable behavior (NaN outputs) for one ScanNet scene. We exclude this scene from all fine-tuning experiments.

\subsection{Baseline Evaluation with Existing 3D LMMs}
\begin{table}[tb]
  \centering
  \caption{Evaluation with Zero-shot 3D LMMs on proposed dataset}
  \begin{tabular}{@{}lcccc@{}}
    \toprule
    Method & visual prompt & mIoU $\uparrow$ & Acc@0.25 $\uparrow$ & Acc@0.50 $\uparrow$ \\
    \midrule
    3D-LLaVA & $\checkmark$ & \textbf{0.0755} & \textbf{0.1167} & \textbf{0.0562}\\
    3D-LLaVA & $\times$ & 0.0720 & 0.1099 & 0.0560\\
    LLaVA-3D & $\checkmark$ & 0.0002 & 0.0002 & 0.0000\\
    LLaVA-3D & $\times$ & 0.0003 & 0.0002 & 0.0000\\
    Reason3D & $\times$ & 0.0321 & 0.0363 & 0.0102\\
    \bottomrule
  \end{tabular}
  \label{tab:zeroshot}
\end{table}

We first compare existing 3D Large Multimodal Models (LMMs) in a zero-shot setting. Specifically, we evaluate 3D-LLaVA~\cite{deng20253dllava}, LLaVA-3D~\cite{zhu2024llava}, and Reason3D~\cite{reason3d}.
For 3D-LLaVA and LLaVA-3D, visual prompts can be provided as additional inputs. We provide an anchor object mask as the prompt for 3D-LLaVA, and a point at the anchor object centroid for LLaVA-3D, which are the closest prompt types supported by each model.

The original LLaVA-3D paper reports results using a model with a Grounding Decoder. As of January 2026, the publicly released implementation does not include this component, so we evaluate the available implementation for reproducibility.

To minimize mismatch between the inference prompt and each model's instruction format, we use prompt templates that closely resemble those used during pretraining:
\begin{itemize}
    \item \textbf{3D-LLaVA:} 
    \texttt{Please output the segmentation mask according to the following description. \textbackslash n the object that is \{relation\} the highlighted \{anchor name\}.}
    \item \textbf{LLaVA-3D:} 
    \texttt{Please output the 3D bounding box of the object(s) that is/are \{relation\} the highlighted \{anchor name\}.}
    \item \textbf{Reason3D:} 
    \texttt{Please segment the object according to the given 3D scene and the description: the object that is \{relation\} the highlighted \{anchor name\}.}
\end{itemize}

For evaluation, we use the dataset split containing 10 viewpoints per scene. We report mIoU as the primary segmentation metric, along with Acc@0.25 and Acc@0.50.
When multiple ground-truth target instances exist, we compute the maximum IoU over all individual instances and the union mask.

Results are summarized in Table~\ref{tab:zeroshot}. Even with visual prompts, 3D-LLaVA achieves very low mIoU (0.0755), which is consistent with the fact that the original model does not take viewpoint as an explicit input. LLaVA-3D performs near zero in this setting, since without a Grounding Decoder the model must directly generate numerical 3D box coordinates. Reason3D does not support visual prompts, which makes anchor specification difficult in scenes that contain multiple instances of the same category. Overall, current 3D LMMs provide only limited baseline performance on viewpoint-dependent referring segmentation.

\subsection{Effect of the Viewpoint Encoder}

\begin{table}[tb]
  \centering
  \caption{Comparison of methods for incorporating viewpoint information into 3D-LLaVA and Reason3D.}
  \resizebox{\columnwidth}{!}{
  \begin{tabular}{@{}lccc@{}}
    \toprule
    Method & mIoU $\uparrow$ & Acc@0.25 $\uparrow$ & Acc@0.50 $\uparrow$\\
    \midrule
    \multicolumn{4}{@{}l}{\textbf{3D-LLaVA}} \\
    Zero-shot & 0.0737 & 0.1140 & 0.0548\\
    LoRA w/o Viewpoint & 0.2960 & 0.4506 & 0.2640\\
    LoRA w/ Text Viewpoint & 0.4457 & 0.6587 & 0.4379 \\
    LoRA w/ Viewpoint Encoder (ours) & \textbf{0.4737} & \textbf{0.6890} & \textbf{0.4760}\\
    \midrule
    \multicolumn{4}{@{}l}{\textbf{Reason3D}} \\
    Zero-shot & 0.0321 & 0.0363 & 0.0102 \\
    Finetuning w/ Viewpoint Encoder & 0.1104 & 0.1749 & 0.0792 \\
    \bottomrule 
  \end{tabular}
  }
  \label{tab:main}
\end{table}

We next study how different ways of providing viewpoint information affect performance. For 3D-LLaVA, we compare: (i) zero-shot evaluation, (ii) LoRA fine-tuning without viewpoint input, (iii) LoRA fine-tuning with viewpoint provided as raw text (a numerical matrix in the prompt), and (iv) LoRA fine-tuning with the proposed viewpoint encoder, which inserts a dedicated viewpoint token.
For training we use the split with 10 viewpoints per scene, and for evaluation we use the split with 500 viewpoints per scene. Unless otherwise noted, we train for 1 epoch with viewpoint encoder hidden dimension 4096, and we set the multitask loss weight to $\lambda=0.1$.

As shown in Table~\ref{tab:main}, LoRA fine-tuning without viewpoint information improves performance over the zero-shot baseline, but remains substantially below the viewpoint-conditioned settings. Providing viewpoint as text yields a large gain, indicating that explicit viewpoint input is important for this task. The proposed viewpoint encoder performs best, suggesting that representing camera pose as a structured token is more effective than supplying raw numerical matrices as text.
We also evaluate fine-tuning Reason3D with the same viewpoint encoder. While the overall performance remains lower than 3D-LLaVA, explicit viewpoint conditioning improves Reason3D compared to its zero-shot baseline.

\subsection{Qualitative Evaluation}
\begin{figure*}[tb]
    \begin{tabular}{cc}
        \begin{minipage}[t]{0.24\hsize}
        \centering
        \includegraphics[keepaspectratio, width=1.0\linewidth]{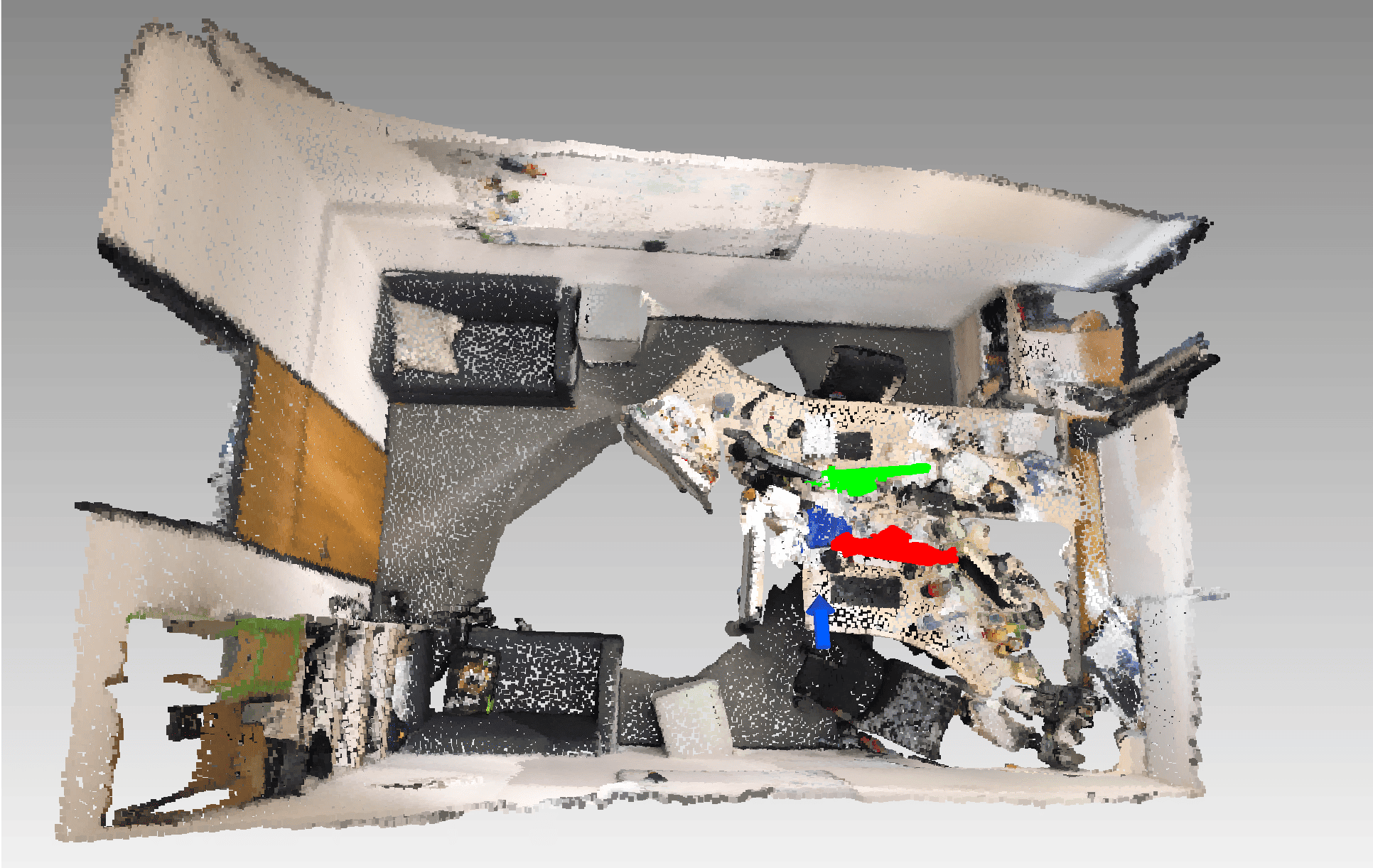}
        \subcaption{Ground Truth}
        \label{fig:gt}
        \end{minipage} 
        \begin{minipage}[t]{0.24\hsize}
        \centering
        \includegraphics[keepaspectratio, width=1.0\linewidth]{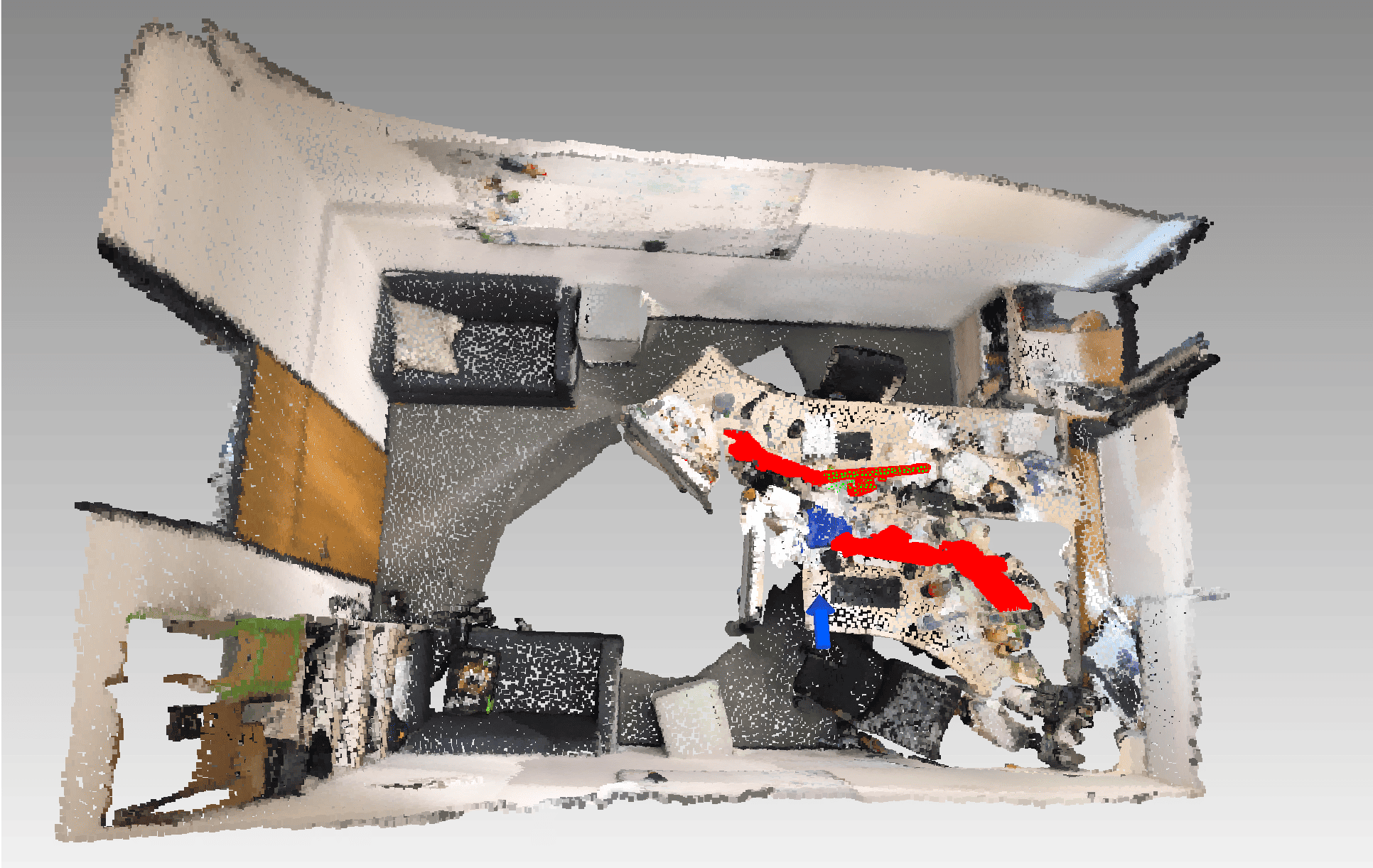}
        \subcaption{Zeroshot 3D-LLaVA}
        \label{fig:zeroshot}
        \end{minipage} 
        
        \begin{minipage}[t]{0.24\hsize}
        \centering
        \includegraphics[keepaspectratio, width=1.0\linewidth]{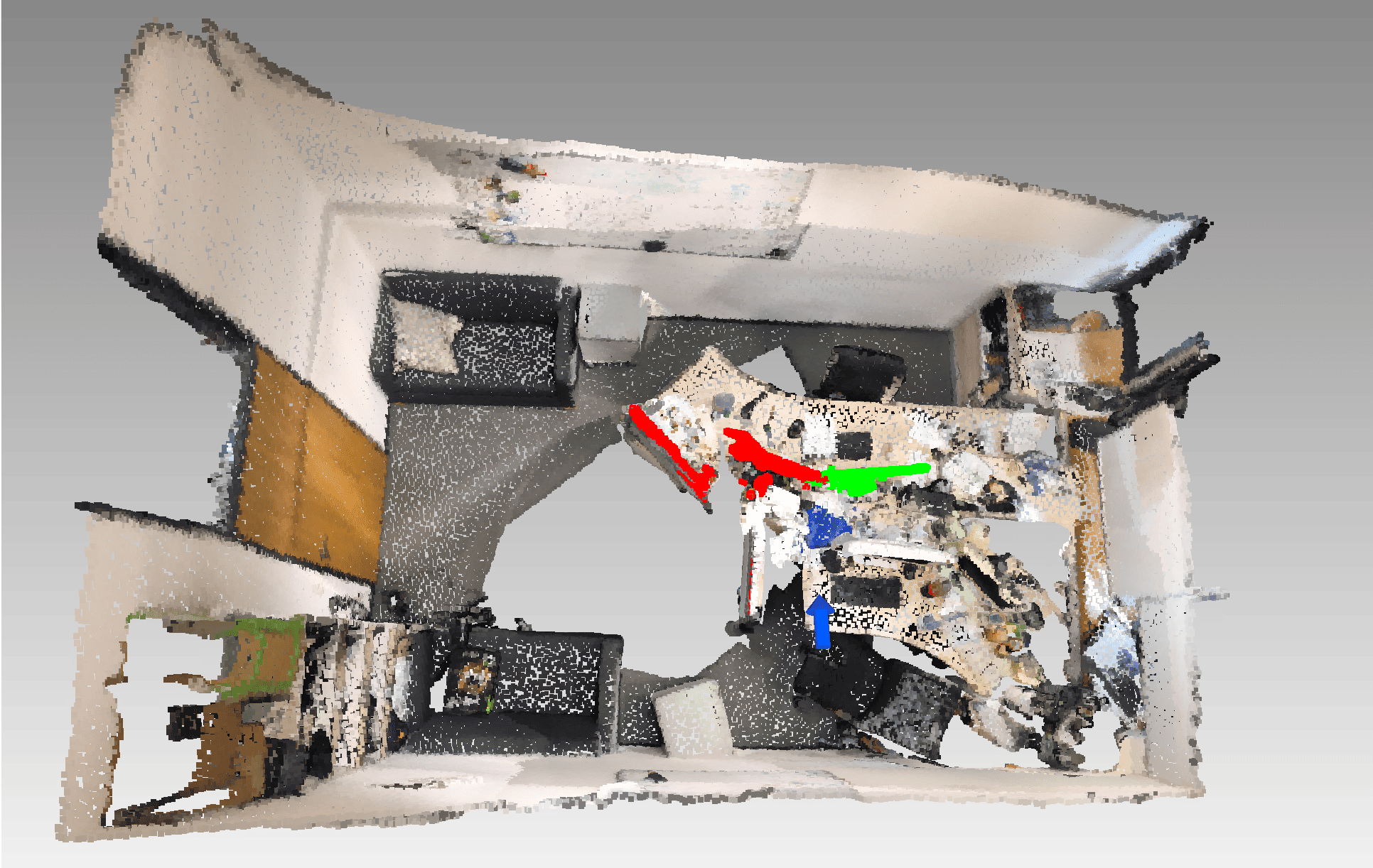}
        \subcaption{Finetuned 3D-LLaVA (w/o viewpoint)}
        \label{fig:wo_viewpoint}
        \end{minipage} 
        \begin{minipage}[t]{0.24\hsize}
        \centering
        \includegraphics[keepaspectratio, width=1.0\linewidth]{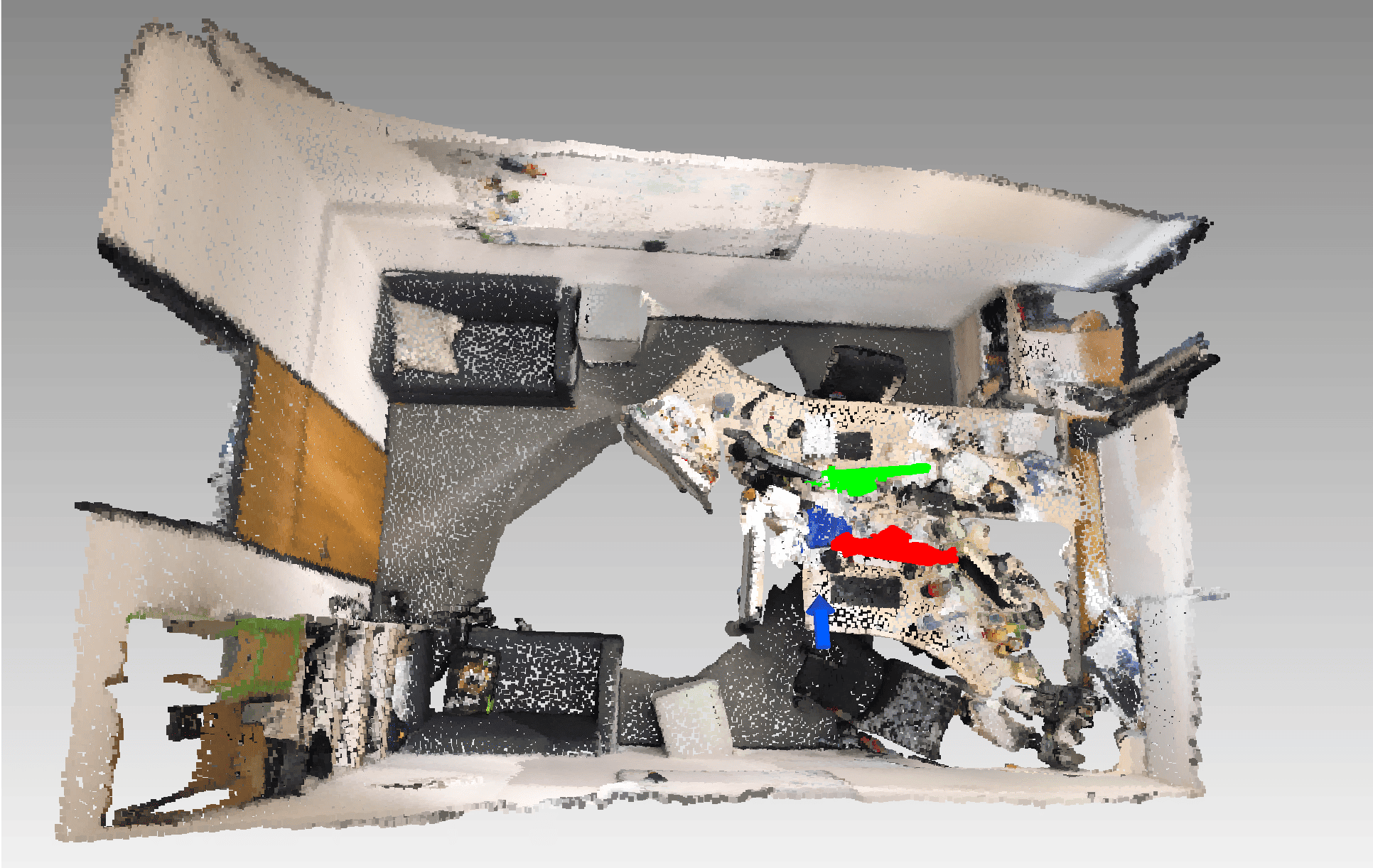}
        \subcaption{Finetuned 3D-LLaVA (w/ viewpoint)}
        \label{fig:ours}
        \end{minipage} 
    \end{tabular}
    \caption{\textbf{Qualitative comparison of referring expression segmentation.}
The green mask indicates the visual prompt, and the blue arrow represents the input viewpoint.
The red mask denotes the ground-truth mask in (a), and the predicted masks generated by each model in (b)–(d). The language prompt used is ``the object that is in front of the highlighted monitor at \texttt{<loc>}, relative to the camera pose \texttt{<viewpoint>}.''}
    \label{fig:qualitative}
\end{figure*}

Figure~\ref{fig:qualitative} shows qualitative comparisons. The zero-shot model often segments the highlighted region or nearby objects rather than the intended referent, suggesting that the prompt is used without reliable viewpoint-dependent interpretation. Fine-tuning without viewpoint information improves overall segmentation quality, but the selected object still often violates the intended relation. In contrast, the proposed method produces masks that better match the target specified by viewpoint-dependent instructions.

\subsection{Ablation}
\begin{table}[tb]
  \centering
  \caption{Effect of the number of training epochs and viewpoint encoder width.}
  \begin{tabular}{@{}llccc@{}}
    \toprule
    Epochs & hidden dim & mIoU $\uparrow$ & Acc@0.25 $\uparrow$ & Acc@0.50 $\uparrow$ \\
    \midrule
    1 & 2048 & 0.4725 & 0.6851 & 0.4777\\
    1 & 4096 & 0.4737 & 0.6890 & 0.4760\\
    1 & 8192 & 0.4700 & 0.6837 & 0.4760 \\
    3 & 4096 & \textbf{0.5085} & \textbf{0.7186} & \textbf{0.5291} \\
    \bottomrule
  \end{tabular}
  \label{tab:epochs_hidden_width}
\end{table}

Table~\ref{tab:epochs_hidden_width} reports ablations on the number of training epochs and the hidden dimension of the viewpoint encoder. Increasing training from 1 to 3 epochs improves mIoU from 0.4737 to 0.5085. For the encoder width, we evaluate hidden dimensions of 2048, 4096, and 8192. The 4096 setting achieves the best performance, while both smaller (2048) and larger (8192) dimensions slightly degrade results. This suggests that smaller or larger viewpoint embeddings do not necessarily improve viewpoint-conditioned grounding under the same training budget.
\section{Limitations}

Although the proposed benchmark enables controlled evaluation of viewpoint-dependent spatial relations, it is currently limited to indoor scenes derived from ScanNet~\cite{dai2017scannet} and a restricted set of spatial relations. Extending the benchmark to more diverse environments and richer linguistic expressions would further improve its applicability.

In addition, while the proposed viewpoint encoder significantly improves performance, challenging cases remain when multiple objects share similar spatial configurations or when target objects occupy small regions of the scene.

\section{Conclusion}

We introduced a viewpoint-aware 3D referring segmentation benchmark that explicitly defines observer-centered spatial relations using geometric criteria, enabling systematic evaluation of viewpoint-dependent spatial reasoning. Experiments showed that existing 3D large multimodal models struggle to interpret such relations when viewpoint information is not explicitly provided. We further proposed a simple viewpoint encoder that incorporates camera pose into a 3D large multimodal model. The proposed approach significantly improves performance on viewpoint-dependent referring segmentation, demonstrating the importance of explicit viewpoint conditioning for spatial reasoning in 3D scene understanding.

More broadly, interpreting viewpoint-dependent spatial relations is essential for robots operating in real-world environments. We hope that the proposed benchmark and viewpoint-aware modeling approach will facilitate future research on spatial language grounding for embodied agents and contribute to more robust perception and interaction in robotics systems.







\bibliographystyle{ieeetr}
\bibliography{references}

\end{document}